\documentclass[letterpaper, 10 pt, conference]{ieeeconf}
\IEEEoverridecommandlockouts
\usepackage{graphics} 
\usepackage{graphicx} 
\usepackage{subcaption} 
\usepackage{amssymb}  
\usepackage{amsmath} 
\usepackage{todonotes}
\usepackage[ruled,vlined]{algorithm2e}
\usepackage{booktabs}

\title{ \LARGE \bf Towards Proprioceptive Terrain Mapping with Quadruped Robots for Exploration in Planetary Permanently Shadowed Regions }

\author{
Alberto Sanchez-Delgado$^{*}$, João Carlos Virgolino Soares, Victor Barasuol, Claudio Semini%
\thanks{This work was supported by the Italian Space Agency (ASI)}
\thanks{All the authors are with the Dynamic Legged Systems lab (DLS), Istituto Italiano di Tecnologia (IIT), Genoa, Italy. \texttt{\{first\_name.first\_surname\}@iit.it}}%
\thanks{*Corresponding author: \texttt{alberto.sanchez@iit.it}.}%
}

\begin{document}
\maketitle
\begin{abstract}
Permanently Shadowed Regions (PSRs) near the lunar poles are of interest for future exploration due to their potential to contain water ice and preserve geological records. Their complex, uneven terrain favors the use of legged robots, which can traverse challenging surfaces while collecting in-situ data, and have proven effective in Earth analogs, including dark caves, when equipped with onboard lighting. While exteroceptive sensors like cameras and lidars can capture terrain geometry and even semantic information, they cannot quantify its physical interaction with the robot—a capability provided by proprioceptive sensing. We propose a terrain mapping framework for quadruped robots, which estimates elevation, foot slippage, energy cost, and stability margins from internal sensing during locomotion. These metrics are incrementally integrated into a multi-layer 2.5D gridmap that reflects terrain interaction from the robot’s perspective. The system is evaluated in a simulator that mimics a lunar environment, using the 21 kg quadruped robot Aliengo, showing consistent mapping performance under lunar gravity and terrain conditions. 
\end{abstract}

\section{Introduction}

The global interest in lunar exploration has brought particular focus to the Moon’s Permanently Shadowed Regions (PSRs), primarily located near the poles. These regions are of significant scientific and strategic interest due to the potential presence of water ice, a critical resource for long-duration missions, and their capacity to preserve geological records~\cite{chien2024,mo2025,arm2023}. Their inaccessibility has made them some of the least understood locations on the lunar surface, despite their high exploration value~\cite{eke2014,dagar2023}.

Orbital missions such as NASA’s Lunar Reconnaissance Orbiter (LRO) and ISRO’s Chandrayaan-2 have studied PSRs through remote sensing, using instruments like LRO’s camera (LROC) and the Lunar Orbiter Laser Altimeter (LOLA) to detect low-reflectance areas and infer terrain profiles~\cite{eke2014}. However, not all of these instruments can operate effectively under the extremely low sunlight incidence angles at the lunar poles. In PSRs, direct illumination is absent, rendering optical observations unreliable or even infeasible~\cite{litaker2025}. This limitation increasingly motivates the use of in-situ robotic systems capable of physically interacting with the terrain. Upcoming potential surface missions, such as NASA’s VIPER rover and the Luna-27 lander, aim to deploy instruments specifically designed for PSRs~\cite{viper2021}.

A trend in planetary exploration is the deployment of multi-robot teams, where different agents assume specialized roles such as terrain scouting, scientific sampling, communications, or logistics~\cite{arm2023}. This architecture improves mission robustness and allows concurrent task execution. In such teams, high-mobility exploratory rovers are essential for generating initial terrain assessments that guide the operation of more sensitive or resource-limited agents~\cite{arm2023,varadharajan2025}. 

Legged robots are particularly suited for this role. Their morphology enables traversal of steep slopes, crater rims, and rocky surfaces by dynamically adjusting contact points and distributing forces across multiple limbs~\cite{Valsecchi2023}. They can adopt diverse locomotion patterns, such as trotting, crawling, or sliding over granular surfaces using their torso on inclined terrain~\cite{barasuol23astra, sanchez25astra}, and use their limbs as active terrain probes to sense compliance, detect slippage, and evaluate stability~\cite{arm2023}.

While exteroceptive sensors such as cameras and lidars can provide rich geometric and semantic terrain descriptions, they cannot quantify
its physical interaction with the robot. Proprioceptive sensing, on the other hand, can fill this gap by quantifying interaction-based metrics critical for mobility and safety.

This work introduces a proprioceptive terrain mapping framework that enables legged rovers to build enriched terrain maps using only internal sensing. These exploratory agents operate in unmapped, visually degraded environments, such as PSRs. Our method estimates elevation, slippage, Cost of Transport (CoT), and stability margins from online proprioceptive feedback. The resulting multi-layer 2.5D gridmap captures not only the terrain's shape but also its physical interaction profile. While relying solely on proprioception, this approach is not intended to replace exteroceptive systems, but to complement them, increasing overall terrain assessment reliability. The generated maps support mission-wide planning by reducing uncertainty over unvisited areas and enabling path comparisons for future agents or instruments.

\subsection{Related Work}
Some works have advanced terrain mapping by integrating multi-modal or semantic layers into elevation representations. Approaches such as MEM~\cite{erniMEM2023}, SEM~\cite{maRealtime2025}, and multi-sensor fusion systems~\cite{ewenYouve2024} combine elevation with contact, material, or semantic data to generate rich 2.5D terrain maps. These methods enable advanced planning and terrain understanding, yet rely on visual or depth information. Additionally, the metrics they encode tend to describe surface appearance or geometry, rather than the physical interaction experienced during traversal.

Other methods focus on terrain classification or traversability estimation from proprioceptive or multi-sensor data. Cai et al.~\cite{caiPIETRA2025}, Elnoor et al.~\cite{elnoor2024}, and Zhang et al.~\cite{zhangTerrain2025} estimate terrain type, roughness, or slippage based on internal measurements. These systems improve local adaptation, but typically lack spatial memory or map-level representations. Their outputs are local and reactive, limiting their use for planning over extended trajectories or sharing information among agents. Some frameworks combine perception and locomotion adaptation. Notably, Ren et al.~\cite{renTOPNav2024}, Miki et al.~\cite{mikiLearning2022}, and Fan et al.~\cite{fanApplication2025} propose systems that integrate terrain awareness into planning or gait selection. These works show robust local control through multi-scale planning or learned locomotion policies, but often assume structured or semi-structured environments and rely on LiDAR data for local terrain estimation. 

Efforts toward planetary navigation often rely on global Digital Elevation Models (DEM) or structured risk models. For example, Santra et al.~\cite{santraRiskAware2024} and Hong et al.~\cite{hongRobotic2022} introduce risk-aware planning and mapping methods that incorporate elevation and illumination variability, while Cheng et al.~\cite{chengVLMEmpowered2025} explore multi-modal policy switching for space robots. However, these approaches assume some form of prior terrain knowledge or visual input, and do not leverage proprioception as a primary information source. Furthermore, they generally target wheeled robots or high-level planning, without encoding the physical demands of traversal into maps. Stability and energy cost estimation are required for safe and efficient navigation in planetary terrain. Traditional stability analysis based on static support polygons~\cite{BiswalDevelopment2021} does not account for inertial effects that dominate in low gravity and steep slopes, common in PSRs. Metrics based on the Gravito-Inertial Acceleration (GIA) vector offer  estimates in these  contexts~\cite{ribeiroDynamicEquilibriumClimbing2020,unoSimulationBased2022}, but prior work applies them only at discrete time steps and not within spatially grounded maps. Similarly, CoT is often treated as a global benchmark~\cite{ribeiroDynamicEquilibriumClimbing2020,Valsecchi2023}, rather than being mapped in relation to terrain features.

Together, these observations suggest that while existing methods have made significant progress in perception and control, they typically do not construct  representations of the terrain grounded in proprioceptive experience. In particular, dynamic stability margins, energetic cost, and slippage are not commonly encoded into maps in a way that supports planning under the degraded sensing conditions typical of PSRs, where physical interaction becomes essential. 

\subsection{Main contributions}

This paper introduces a proprioceptive terrain mapping framework designed for legged planetary robots operating in visually degraded environments such as lunar PSRs. The proposed approach seeks to address current limitations in terrain understanding by capturing key aspects of the robot’s physical interaction with the environment during locomotion. The framework is intended to support terrain-aware planning and coordination in unstructured settings while complementing exteroceptive sensing. The main contributions are:

\begin{itemize}
\item \textbf{Proprioceptive terrain mapping:} We propose a method to build multi-layer 2.5D maps from proprioceptive signals collected online during locomotion that encode elevation, foot slippage, CoT, and  stability margins. These layers offer a physically grounded representation of terrain interaction. The system is validated replicating PSR conditions in OmniLRS \cite{richardOmniLRS2024}.

\item \textbf{Robot-centered terrain assessment:} The resulting map reflects how the terrain behaves from the robot’s own physical experience (how it feels to traverse it), enabling the evaluation and comparison of candidate paths based on effort, stability, and ground interaction, rather than geometric descriptors alone.

\item \textbf{GIA-based spatial stability mapping:} We construct spatial maps of Gravito-Inertial stability margins under reduced gravity. This extends their previous use in climbing ~\cite{unoSimulationBased2022} and reinforcement learning~\cite{Valsecchi2023}, providing localized indicators of risk. 
\end{itemize}


\section{ PROPRIOCEPTIVE TERRAIN MAPPING FRAMEWORK}
\subsection{Overview of the Framework} 

This section introduces our proprioceptive terrain mapping framework as a strategy to construct actionable and persistent knowledge of the environment from the robot's own physical interaction with it. 

A quadruped rover walking through unknown regions, collects mechanical and dynamic cues from its interaction with the ground. These cues encode terrain-level properties that are both robot-relevant and mission-relevant. By continuously estimating and spatially registering these signals, our framework builds a multi-modal gridmap that reflects not just the terrain geometry, but how it feels from the robot’s perspective to traverse it.

\begin{figure*}[htbp]
    \centering
    \includegraphics[width=0.8\linewidth]{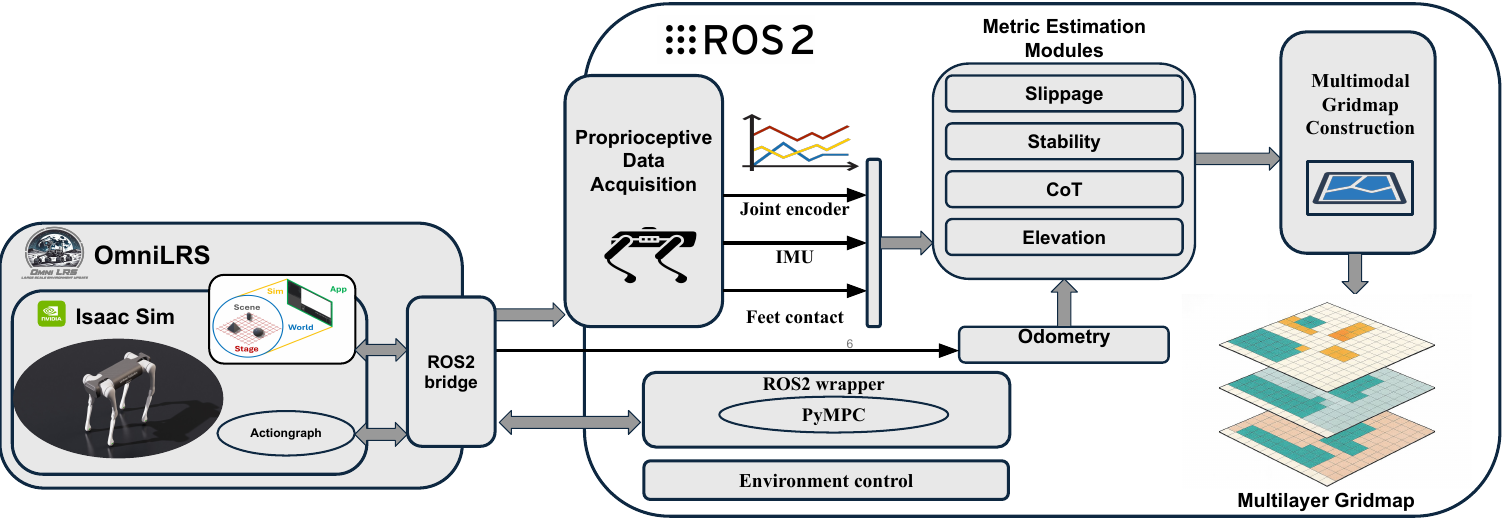}
    \caption{Diagram of the proposed proprioceptive terrain mapping framework. The system processes raw proprioceptive signals to estimate terrain metrics, which are then integrated into a multi-layer gridmap.}
    \label{fig:framework_diagram}
\end{figure*}

The system focuses on four proprioceptively estimated terrain metrics: foot slippage, which reveals regions prone to slipping; CoT, which reflects energetic demands; stability margins, which quantify risk of tumbling; and elevation, which provides geometric context. Together, these metrics capture additional aspects of the robot's interaction with the terrain and enable a multi-dimensional characterization of the environment.

As shown in Fig.~\ref{fig:framework_diagram}, the proposed framework is structured as a modular pipeline composed of three main stages:

\begin{enumerate}
    \item \textbf{Proprioceptive Data Acquisition.} During locomotion, the robot collects raw proprioceptive signals including joint torques, joint velocities, foot contact states, and base accelerations. These internal measurements are continuously streamed and serve as the input to the estimation process.

    \item \textbf{Metric Estimation Modules.} This stage processes the acquired signals to estimate terrain-relevant metrics: foot slippage, CoT, stability margins, and elevation. Each metric is computed independently and aligned with the robot’s motion, producing localized descriptors of the terrain interaction.

    \item \textbf{Proprioceptive Gridmap Construction.} The estimated metrics are spatially assigned to a fixed-resolution gridmap based on foot positions and robot pose. Repeated visits to the same cell are aggregated incrementally to smooth temporal noise. The resulting map includes layers such as: proprioceptive elevation, slippage, CoT and GIA stability updated online and referenced to the robot’s initial location. This terrain memory supports path comparison, and information sharing among agents.
\end{enumerate}

Finally, the maps generated by the exploratory robot can be shared with other agents, robotic or human, to support collaborative planning. By encoding the perceived effort, risk, and instability along different paths, it gives another tool to collective situational awareness in lunar missions.

\subsection{Proprioceptive Estimation Modules}

This section describes the proprioceptive terrain metrics computed by the framework during locomotion. Each metric is estimated online from internal sensing signals and captures a specific aspect of the terrain's physical interaction properties. Together, they form the layers of the gridmap.

\subsubsection{\textbf{Slip Detection}} 

The slip detection method employed in this work is based on the approach proposed by Nisticò et al.~\cite{nisticoSlip2022}. It relies solely on proprioceptive signals and computes slip indicators in the robot’s base frame, thereby avoiding drift accumulation associated with world-frame estimates. For each foot in stance phase, two quantities are evaluated: the velocity deviation $\Delta V$ and the positional discrepancy $\Delta P$. The velocity deviation is computed as:
\begin{equation}
\Delta V = \sqrt{ \sum_{i = x,y,z} \left( \frac{\dot{x}^{\text{des}}_{f,i} - \dot{x}_{f,i}}{|\dot{x}^{\text{des}}_{f,i}| + h} \right)^2 }
\label{eq:delta_v}
\end{equation}
where $\dot{x}^{\text{des}}_{f,i}$ and $\dot{x}_{f,i}$ are the desired and actual foot velocities along axis $i$ in the base frame, and $h$ is a small positive constant to ensure numerical stability. The positional discrepancy is defined as:

\begin{equation}
\Delta P = \left| \| \mathbf{x}^{\text{des}}_f \| - \| \mathbf{x}_f \| \right|
\label{eq:delta_p}
\end{equation}
with $\mathbf{x}^{\text{des}}_f$ and $\mathbf{x}_f$ denoting the desired and actual foot positions (computed using leg kinematics with the Pinocchio library), respectively.

A slip event is flagged when both quantities exceed predefined thresholds, as follows:
\begin{equation}
\beta =
\begin{cases}
1 & \text{if } \Delta V > \varepsilon_v \text{ and } \Delta P > \varepsilon_p \\
0 & \text{otherwise}
\end{cases}
\label{eq:slip_flag}
\end{equation}
Here, $\varepsilon_v$ is a threshold (percentile-based on recent $\Delta V$), while $\varepsilon_p$ is a fixed positional threshold tuned empirically. \newline

\subsubsection{\textbf{Cost of Transport}}
To evaluate energetic efficiency in the legged system we made use of the Cost of Transport (see~(\ref{eq:cot})), defined as the energy required to move a unit weight over a unit distance. As a dimensionless quantity, it enables  comparisons across different gaits, robots, and terrain types. We compute CoT as:

\begin{equation}
    \text{CoT} = \frac{E}{m g d}
    \label{eq:cot}
\end{equation}

where $E$ is the mechanical energy consumed (see~(\ref{eq:energy})), $m$ is the mass of the robot, $g$ is the gravitational acceleration (set to $1.62~\text{m/s}^2$ for lunar conditions), and $d$ is the distance traveled.

The energy $E$ is estimated from joint-level proprioceptive signals, specifically the torques $\tau_i(t)$ and angular velocities $\dot{\theta}_i(t)$ of each motor $i$:

\begin{equation}
    E = \int_{t_0}^{t_f} \sum_{i=1}^{N} |\tau_i(t) \cdot \dot{\theta}_i(t)| \, dt
    \label{eq:energy}
\end{equation}

where $N$ is the total number of actuated joints. The absolute value ensures that all power contributions are positive, regardless of joint direction. This formulation captures mechanical energy expenditure, which is a valid proxy in simulation where thermal and electrical losses are assumed negligible or constant.

The computed CoT values are incrementally assigned to terrain map cells during locomotion, enabling the construction of energy-aware terrain layer.

\subsubsection{\textbf{Stability Margins}}
Under a planetary scenario with low-gravity and steep environment conditions, traditional static stability models based on center-of-mass projections may be unreliable, as they neglect inertial effects that become significant during motion. To address this, we adopt the concept of \textit{Gravito-Inertial Acceleration} (GIA)~\cite{unoSimulationBased2022}, which accounts for both gravitational and motion-induced forces, as follows:

\begin{equation}
    \mathbf{a}_{\text{GIA}} = \mathbf{g} - \frac{1}{m} \sum_{i=1}^{N} m_i \mathbf{a}_i
    \label{eq:gia}
\end{equation}

Here, \( \mathbf{g} \) is the gravitational acceleration vector, \( m \) is the robot’s total mass, and \( m_i \), \( \mathbf{a}_i \) are the mass and acceleration of segment \( i \). The robot is considered stable if the GIA vector remains inside the \textit{stability polyhedron} formed by the current support contacts.

We quantify stability using two complementary margins:

\begin{itemize}
    \item \textbf{Gravito-Inertial Inclination Margin (GIIM)}: the smallest angle between the GIA vector and the outward normals of the polyhedron’s faces:
    \begin{equation}
        \theta_{\text{margin}} = \min_{\text{faces}} \left[ \arccos\left( \frac{\mathbf{n}_{ab} \cdot \mathbf{a}_{\text{GIA}}}{\|\mathbf{n}_{ab}\| \|\mathbf{a}_{\text{GIA}}\|} \right) \right] - \frac{\pi}{2}
        \label{eq:giim}
    \end{equation}

    \item \textbf{Gravito-Inertial Acceleration Margin (GIAM)}: the minimum acceleration (projected) of the GIA vector from any face of the stability polyhedron:
    \begin{equation}
        a_{\text{margin}} = \min_{\text{faces}} \left[ -\frac{\mathbf{n}_{ab} \cdot \mathbf{a}_{\text{GIA}}}{\|\mathbf{n}_{ab}\|} \right]
        \label{eq:giam}
    \end{equation}
\end{itemize}

These margins provide physical indicators of stability and the risk of tumbling over. They indicate how much additional acceleration and inclination the system can tolerate before tipping occurs. Their inclusion is relevant in PSRs, where steep slopes and uneven terrain are common, and where proprioception may be used as a complement to evaluate risk of tumbling.

\subsubsection{\textbf{Elevation Estimation}}
Whenever a stance phase is detected without slippage, the vertical position of the contacting foot, expressed in the world frame, is used to estimate the local terrain elevation. This value is computed by transforming the foot position from the base frame using the estimated base pose:

\begin{equation}
    \mathbf{p}^{\text{world}}_f = \mathbf{T}^{\text{world}}_{\text{base}} \cdot \mathbf{p}^{\text{base}}_f
    \label{eq:foot_world}
\end{equation}

Here, \( \mathbf{p}^{\text{base}}_f \) is the foot position in the robot’s base frame (obtained from forward kinematics), and \( \mathbf{T}^{\text{world}}_{\text{base}} \) is the transformation matrix representing the base pose in the world frame.

The terrain height at the contact location is then given by the vertical coordinate of the transformed foot position as follows:

\begin{equation}
    h = \left[ \mathbf{p}^{\text{world}}_f \right]_z
    \label{eq:elevation}
\end{equation}

When multiple feet contact the same terrain cell over time, elevation values are incrementally averaged to smooth noise and improve robustness.

\subsection{Proprioceptive Terrain Mapping and Multimodal Data Association}

This section describes how proprioceptively estimated metrics are incrementally aggregated into a multi-layer gridmap. Each layer of the map represents a specific terrain property relevant to mobility, and all are spatially anchored to a fixed local reference frame.

\subsubsection{Gridmap Structure and Spatial Assignment}

The terrain is represented as a two-dimensional grid $\mathcal{G} \in \mathbb{R}^{N_x \times N_y}$, where each cell $\mathcal{G}_{i,j}$ encodes a set of proprioceptive quantities collected at position $(x_i, y_j)$. The grid resolution $r$ defines the cell width in meters and a world-aligned coordinate system centered at the robot’s initial pose provides the global reference frame for all updates.

Two spatial referencing strategies are used depending on the nature of the signal: For contact-based metrics (elevation,  slippage), the Cartesian position of each foot during stance is projected to the grid. Meanwhile, for body-centered metrics (CoT, GIIM, GIAM), the grid cell is selected based on the projection of the robot’s CoM over the plane. The indexing from a continuous position $(x, y)$ to grid coordinates $(i, j)$ is computed as follows:

\begin{equation}
i = \left\lfloor \frac{y - y_0}{r} + \frac{N_y}{2} \right\rfloor,\quad
j = \left\lfloor \frac{x - x_0}{r} + \frac{N_x}{2} \right\rfloor
\label{eq:indexing}
\end{equation}

where $(x_0, y_0)$ denotes the origin of the grid. This indexing ensures that the robot’s local surroundings are symmetrically represented around its starting location.

\subsubsection{Cell-Level Data Integration}

Each proprioceptive metric $\phi_k$ is mapped into a corresponding layer $\mathcal{L}_k$ of the grid. When multiple observations fall within the same cell $(i, j)$, either due to multiple legs or revisits, the value is updated incrementally to reduce the impact of noise.

\begin{equation}
\phi_k^{(i,j)} \leftarrow \frac{n \cdot \phi_k^{(i,j)} + \hat{\phi}_k}{n + 1}
\label{eq:update}
\end{equation}

where $\hat{\phi}_k$ is the current observation and $n$ is the number of previous updates for that cell. This formulation assumes all measurements are equally weighted and prioritizes convergence over immediate accuracy. Slip detection is used to guard updates for elevation and GIIM/GIAM metrics, ensuring that only valid contacts contribute to the map.

For binary events such as slippage, a discrete event counter is maintained per cell. Let $\beta_f \in \{0,1\}$ indicate a slip event for foot $f$. Then, the slip event count at cell $(i,j)$ is incremented as follows:

\begin{equation}
S_{i,j} \leftarrow S_{i,j} + \sum_{f=1}^{4} \mathbf{1}_{[c_f \land \beta_f]} \cdot \delta(i_f, j_f, i, j)
\label{eq:slip_event}
\end{equation}

where $c_f$ indicates stance contact, and $\delta(\cdot)$ is the Kronecker delta that returns 1 if the position of foot $f$ maps to cell $(i,j)$.

To improve map coherence, the layers undergo optional spatial smoothing. Here, the layers are post-processed using a Gaussian filter $S_{i,j}^{\text{smooth}}$  to highlight consistent patterns and reduce noise. 

\begin{equation}
S_{i,j}^{\text{smooth}} = \sum_{u,v} S_{u,v} \cdot \mathcal{N}\left((i-u, j-v);\, 0, \sigma^2\right)
\label{eq:smoothing}
\end{equation}

with $\sigma$ controlling the degree of spatial smoothing

\subsubsection{Map Representation and Layers}

The gridmap comprises the following layers, each encoding a specific physical interaction metric: Elevation, Slippage,  CoT, GIIM, GIAM.

\section{IMPLEMENTATION AND EVALUATION OF THE SYSTEM}
\subsection{Simulation Environment and Robotic System} 
The experiments were conducted using NVIDIA Isaac Sim and the lunar simulator OmniLRS \cite{richardOmniLRS2024}. The simulator offers integration with ROS 2 and enables robot-sensor interaction under realistic lighting and gravitational conditions. 

\begin{figure}[t]
    \centering
    \includegraphics[width=\linewidth]{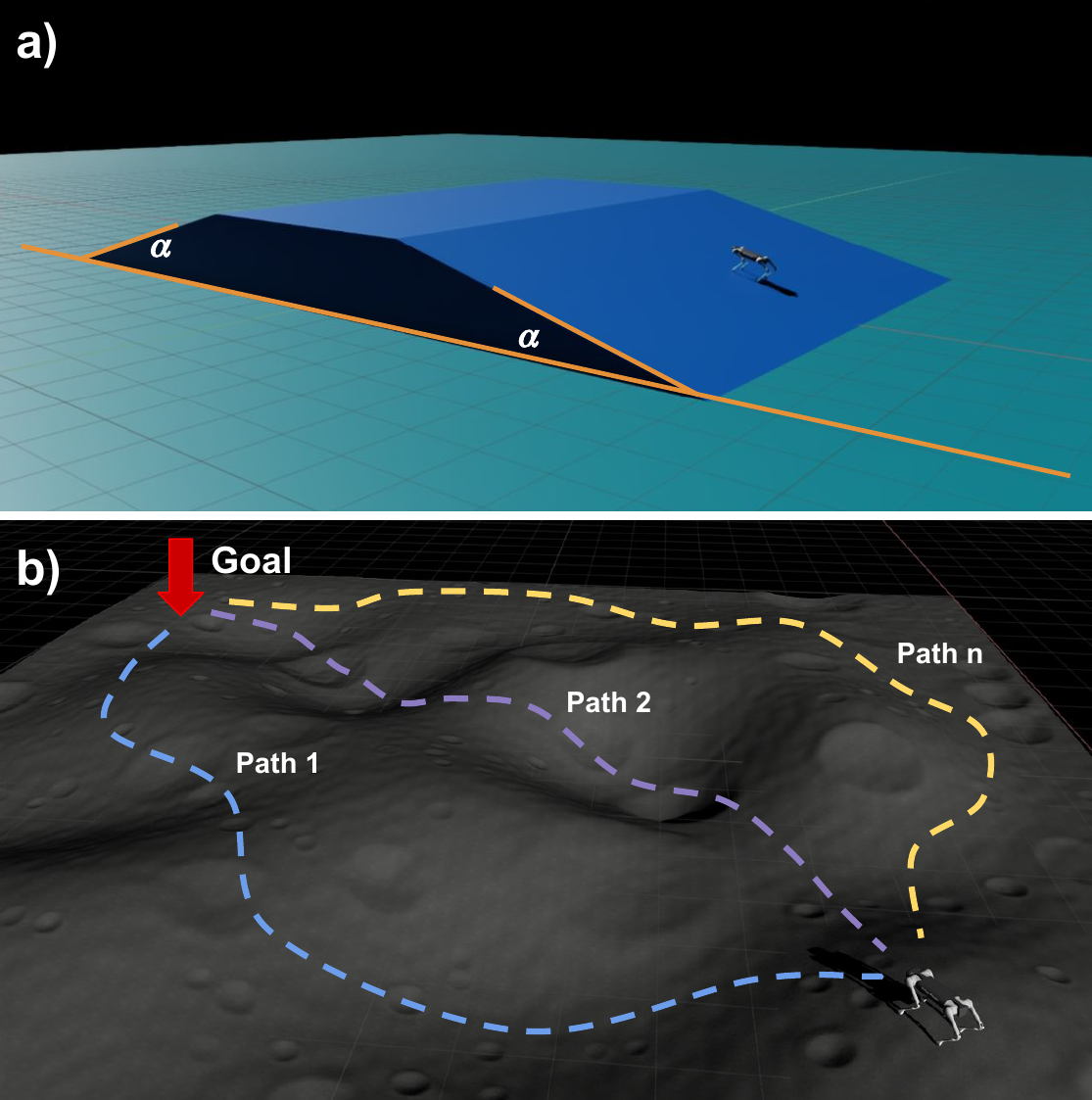}
    \caption{Simulation environments and robot. a) Symmetric ramps of variable inclination angle $\alpha$. b) Lunar terrain in OmniLRS with different curved trajectories and robot sensors.}
    \label{fig:simulation_scenarios}
\end{figure}

In the simulation, the robot is equipped with a suite of proprioceptive sensors as shown in Fig.~\ref{fig:simulation_scenarios}b. These sensors include joint encoders, IMU and  foot contact sensors. No exteroceptive sensor were used in this study.

Low-level locomotion control uses Quadruped-PyMPC, an open-source MPC framework from IIT-DLS lab \footnote{https://github.com/iit-DLSLab/Quadruped-PyMPC} \cite{turrisi2024sampling}. It computes optimal ground reaction forces over a receding horizon to track base motion under dynamics and friction constraints. 

\subsection{Simulation Scenarios and Evaluation Setup}\label{sec:scenarios}
To evaluate the proposed proprioceptive terrain mapping framework, we designed two complementary simulation scenarios under lunar gravity conditions ($g_{\mathrm{moon}} \approx 1.62 \ \mathrm{m/s^2}$), each targeting specific aspects of the system: consistency of metric estimation and comparative path evaluation. Figure~\ref{fig:simulation_scenarios} illustrates the simulated environments.

The first scenario, shown in Fig.~\ref{fig:simulation_scenarios}a, was implemented in Isaac Sim 4.2 and features a structured testbed composed of two ramps and a flat central platform. The ascending and descending ramps share the same inclination angle $\alpha$, which was varied across experiments (5°, 10°, 15°, and 20°). This environment was designed to assess how proprioceptive metrics respond to known changes in terrain geometry under controlled conditions.

The second scenario, depicted in Fig.~\ref{fig:simulation_scenarios}b, consists of a 20~m $\times$ 20~m unstructured lunar-like terrain simulated in OmniLRS. The surface includes elevation variations of up to ±1.5~m, craters up to 2.5~m wide and 1.0~m deep, and scattered rocks. In this case, the robot executed multiple trajectories between fixed start and goal points across the environment. The aim was to evaluate the resulting maps across different traversals and assess how proprioceptive data can inform path selection in the presence of complex terrain variability.

\subsubsection{Experimental Protocol}
All experiments were conducted at a commanded velocity of 0.2~m/s. To ensure consistent motion across terrain variations, a higher-level velocity closed-loop controller was implemented on top of the existing Model Predictive Controller (MPC), which otherwise tends to slow down in response to terrain roughness. For this reason, speeds higher than 0.2~m/s caused repeated falls during slope traversal, making this value the maximum feasible for robust evaluation.

In the first scenario, the robot executed four repeated traversals per slope angle, each time starting from the same initial location. It followed a straight-line trajectory across three ramp segments: initial ascending section, flat platform section and descending section, forming a continuous elevation profile. Between runs, the proprioceptive maps were reset to prevent inter-trial data accumulation, while time series and gridmaps were aggregated offline for subsequent analysis. This first set of experiments aimed to evaluate whether the proposed proprioceptive metrics respond consistently to changes in terrain inclination, and to identify expected fluctuations near geometric transitions, specifically at ramp entries and exits, where ground-robot interactions vary.

A second set of ramp experiments was conducted to investigate how proprioceptive stability margins and energy expenditure (CoT) vary between uphill and downhill locomotion on identical slopes. While the robot always started from the same side of the testbed, both traversal directions were evaluated independently using symmetric ramp segments. The aim was to identify whether these metrics exhibit coherent trends across opposite directions of motion in relation to terrain inclination. For each fixed inclination angle, CoT, GIIM, and GIAM were averaged over their corresponding gridmap layers to obtain representative values for the entire trajectory.

In the second scenario, three additional trajectories were tested. These trajectories were planned from corner to corner following a straight line. One in direct route and the remaining two reaching intermediate landmarks  These paths were selected to provide broad terrain coverage and generate diverse proprioceptive maps for comparison. 

\section{RESULTS}
\subsection{Controlled Ramps: Proprioceptive Map with Geometric Transitions}

\begin{figure}[t]
    \centering
    \includegraphics[width=0.45\textwidth]{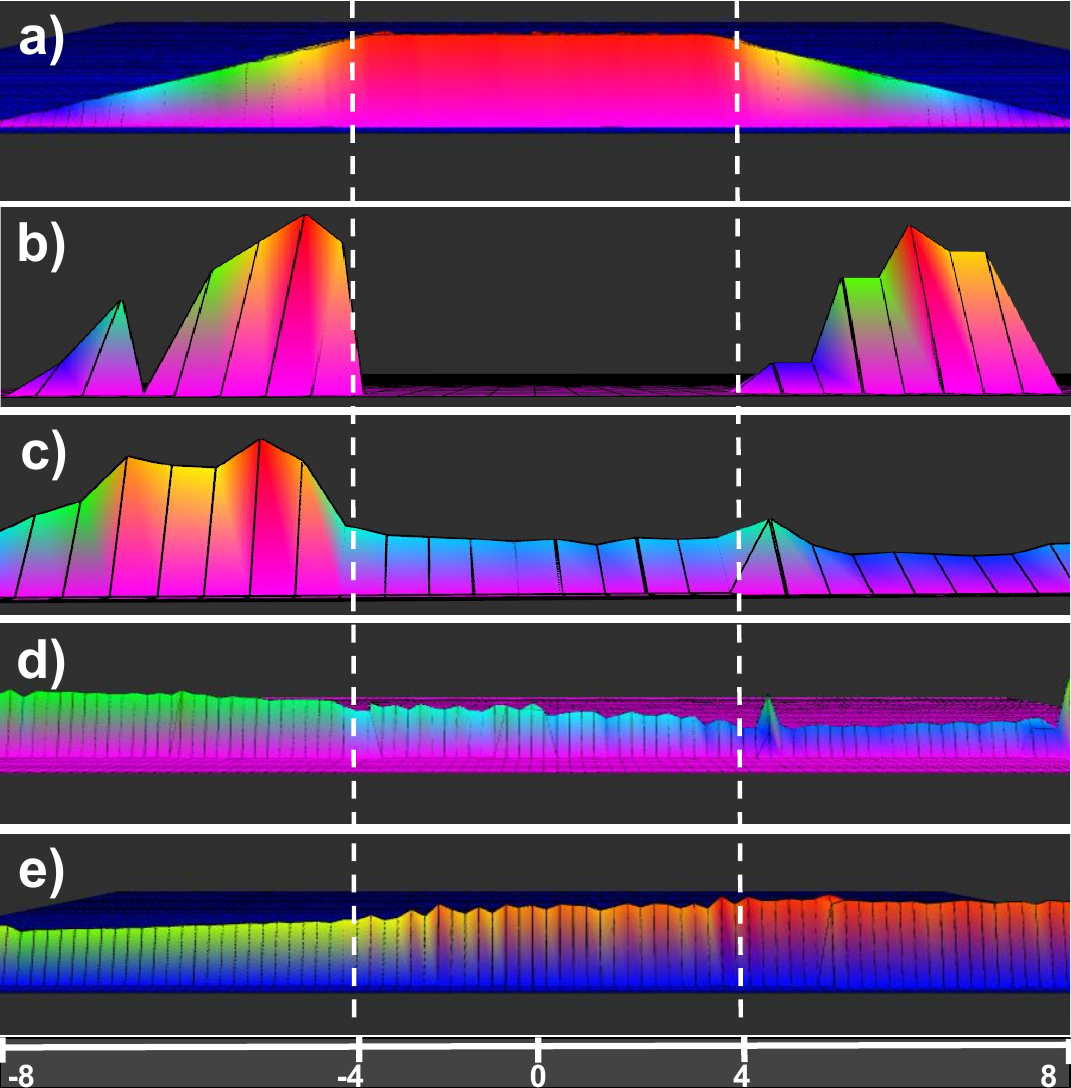}
    \caption{Layers of the proprioceptive terrain map collected in the structured ramp scenario. Each row shows: (a) Elevation, (b) Slippage detection, (c) CoT, (d) GIIM, (e) GIAM. Dashed vertical lines indicate transitions between slopes.}
    \vspace{-0.5cm}
    \label{fig:ramps_results}
\end{figure}

To evaluate the spatial consistency and internal validity of the proposed proprioceptive terrain mapping framework, we first tested the system in the first scenario according to the first set of experiments described in section \ref{sec:scenarios}. The resulting proprioceptive map layers, displayed in lateral view aligned with the walking direction, are presented in Fig.~\ref{fig:ramps_results}. A colormap encodes amplitude values, with cooler colors representing lower amplitudes and warmer colors higher ones. The scale shows the distance between transitions in meters.

The elevation map shown in Fig.~\ref{fig:ramps_results}a captures the terrain geometry with spatial coherence. This map was verified against the known ramp geometry, yielding a Root Mean Square Error (RMSE) of 0.0876 m at a grid resolution of 0.4 m. The three terrain segments: uphill, flat, and downhill, are distinctly reconstructed, and clearly related with the expected topography. Dashed vertical lines indicate the transitions between segments and serve as reference markers across the remaining metric layers.

In the slippage layer (Fig.~\ref{fig:ramps_results}b), slip events are concentrated along the inclined segments of the terrain, with both the ascending and descending ramps exhibiting clear clusters of activity. The flat central section shows negligible slippage, indicating consistent foothold and stable contact dynamics. The concentration on the slopes suggests that increased tangential forces and varying normal loads during inclination transitions remain the dominant contributors to slip occurrence. Quantitative verification using simulator ground-truth foot poses confirmed that detected slip events corresponded to actual displacements exceeding 0.02 m.

The energy cost layer in Fig.~\ref{fig:ramps_results}c also shows a variation pattern related to different slopes. The ascending ramp exhibits the highest CoT, followed by the flat section with intermediate values, and the lowest CoT values are observed in the descending ramp. Although it seems intuitive to extrapolate this trend to angles with greater inclination than those of the current ramp, a broader analysis presented in the following section indicates that this relationship does not hold for them.

Fig.~\ref{fig:ramps_results}d and Fig.~\ref{fig:ramps_results}e depict the spatial distribution of GIIM and GIAM, respectively. The GIIM layer decreases progressively from the ascending to the descending ramp, indicating reduced angular stability as terrain inclination increases. This can be explained by the relative direction of gravity and base acceleration: during uphill locomotion, these vectors oppose each other, increasing the stability margin; while in downhill locomotion, they align, reducing it. In contrast, GIAM exhibits the opposite trend, with higher values on the downhill side and lower values uphill. This inversion is expected within the moderate slope range used in this experiment. As will be shown in the next section, both GIIM and GIAM tend to decrease consistently at higher inclinations.

\subsection{Stability and CoT as a Function of Slope}

To investigate how proprioceptive stability margins and energy expenditure vary as a function of different inclinations, a second set of experiments was conducted on symmetric ramps with angles ranging from $-20^\circ$ to $+20^\circ$. The procedure followed the protocol described in Section~III.C, where the robot executed repeated straight-line traversals over ramp segments with different fixed slope angles. For each inclination, the metrics were recorded along the trajectory and averaged to produce representative values per angle. 

\begin{figure}[t]
    \centering
    \includegraphics[width=0.4\textwidth]{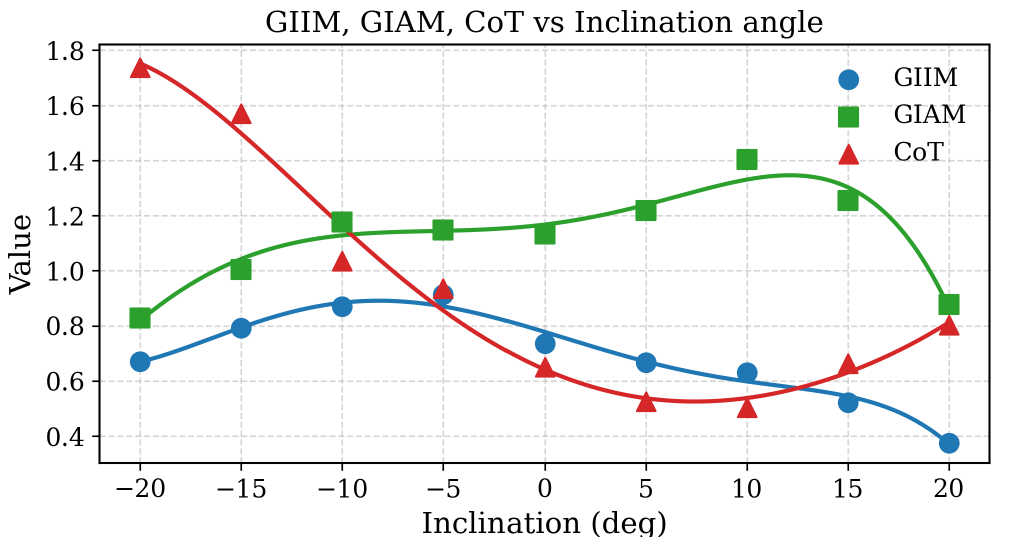}
    \caption{Variation of proprioceptive metrics as a function of ramp inclination. The plot shows: CoT (red triangles), GIIM (blue circles) and GIAM (green squares)}
    \label{fig:cot_stability_inclination}
\end{figure}

Fig.~\ref{fig:cot_stability_inclination} displays the variation of three proprioceptive metrics as a function of terrain inclination: CoT (red triangles), GIIM (blue circles), and GIAM (green squares). The horizontal axis corresponds to the ramp inclination in degrees, while the vertical axis reports the magnitude of each metric in its respective unit: dimensionless for CoT, radians for GIIM, and meters per second squared for GIAM. Fourth-degree polynomial fits were applied to each metric to highlight their trends across the slope range and to support interpolation between sampled angles.

The CoT curve exhibits a general increase toward both ends of the slope range. This trend suggests that energy expenditure rises as the terrain becomes steeper, whether ascending or descending. In the intermediate range between $-15^\circ$ and $+15^\circ$, the CoT values show a minimum near the $5^\circ$ to $10^\circ$ descending slopes. Within this region, gravitational forces contribute to forward motion without requiring substantial corrective effort, which may explain the observed reduction in energetic cost.

In contrast, both stability margins decrease at higher inclinations. For GIIM, the highest values are observed on moderate uphill segments, particularly near $-10^\circ$, and decrease progressively toward both ends of the slope range. This behavior aligns with the interpretation that uphill walking benefits from opposing gravitational and inertial directions, while downhill motion combines these effects, reducing stability. The GIAM metric shows an inverse pattern in the central range: it increases between $-15^\circ$ and $+15^\circ$, reaching higher values on descending slopes compared to flat or ascending terrain. Beyond this range, GIAM also declines, indicating reduced acceleration margins under more extreme inclinations.

\subsection{Proprioceptive Terrain Mapping}
\begin{figure*}[t]
    \centering
    \includegraphics[width=0.75\textwidth]{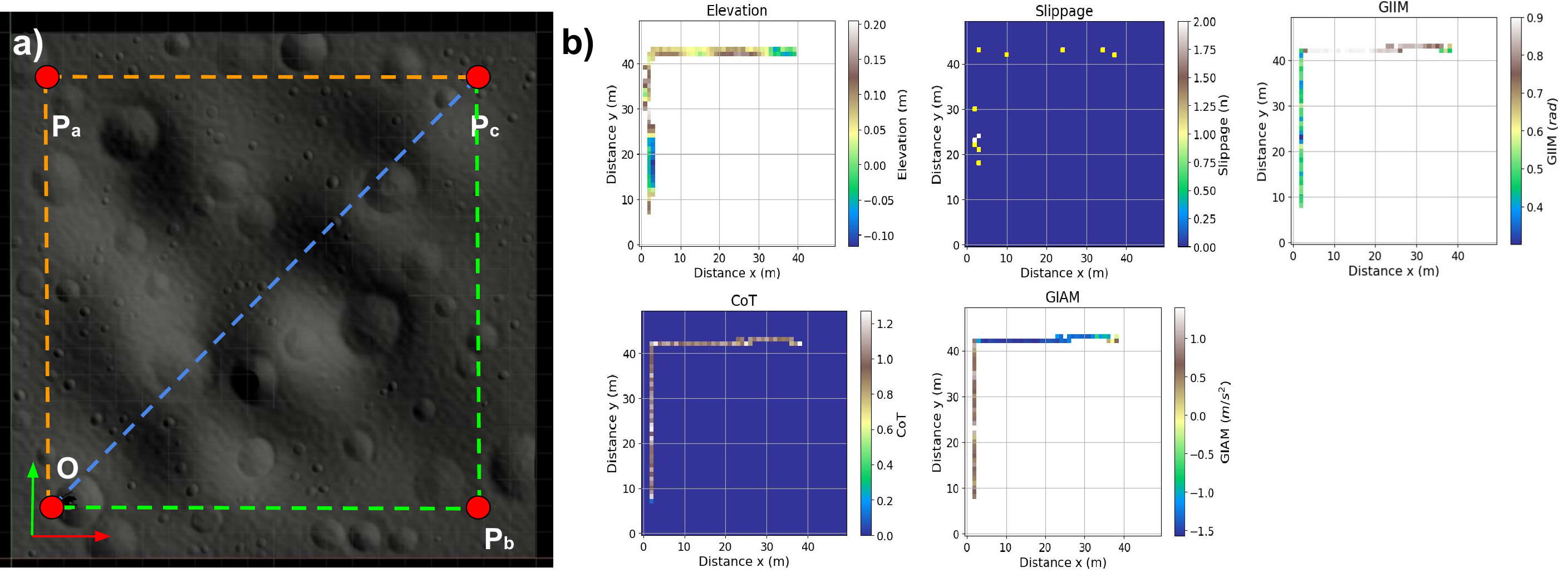}
    \caption{Terrain mapping in the OmniLRS environment. a) Lunar terrain with goal labels. b) Proprioceptive maps}
    \label{fig:mapping_omnilrs}
\end{figure*}

Experiments over this proprioceptive terrain mapping were performed in the second scenario as previously described in section \ref{sec:scenarios}, and using points $P_o$, $P_a$ and $P_b$ as landmarks see Fig.~\ref{fig:mapping_omnilrs}. 

Fig.~\ref{fig:mapping_omnilrs} shows a) the terrain layout depicting the terrain and trajectory overlaid in a top-down view and b) the proprioceptive maps generated during the second trajectory. This example is representative of the mapping behavior observed across all trajectories. The generated maps allow for the identification of terrain features from proprioceptive feedback alone. Changes in elevation and slope are visible as variations in energy cost and slip density. In particular, areas with sustained inclines or uneven topography are associated with increased CoT and higher slip event concentrations. The gridmaps enable localizing these effects and comparing how different paths interact with terrain features.

Table~\ref{tab:map_comparison} summarizes four global metrics extracted from the gridmaps of the three trajectories: total number of slip events, mean CoT, and average values of the GIIM and GIAM. While trajectory 2 shows lower total energy cost and fewer slip events compared to the others, and similar values of GIIM and GIAM, the corresponding map also reveals a localized region of high slippage density and reduced GIIM at the beginning of the path. This zone corresponds to a sustained slope segment, which likely contributes to the concentration of slip events observed in that area. These metrics could indicate traversability by integrating energetic cost, stability, and ground interaction quality. Terrains with low CoT and slippage, and high GIIM/GIAM values, are expected to enable efficient and robust locomotion.

\begin{table}[t]
    \centering
    \caption{Comparison of proprioceptive metrics for three trajectories in OmniLRS terrain}
    \label{tab:map_comparison}
    \resizebox{\columnwidth}{!}{%
    \begin{tabular}{lccc}
        \toprule
        \textbf{Metric} & \textbf{Trajectory 1} & \textbf{Trajectory 2} & \textbf{Trajectory 3} \\
        \midrule
        Total Slippage & 21 & 13 & 15 \\
        Overall CoT & 1.35 & 1.12 & 1.28 \\
        Avg. GIIM & 0.59 & 0.67 & 0.71 \\
        Avg. GIAM & 0.92 & 0.71 & 0.78 \\
        \bottomrule
    \end{tabular}%
    }
\end{table}

\section{LIMITATIONS}
The current framework was evaluated only in simulation, and while the observed spatial patterns are physically plausible, their validity has not yet been experimentally confirmed and may partially reflect controller or simulation artifacts. Testing in lunar analog sites will be essential to assess performance under real-world noise and unmodeled terrain effects. All simulations were performed on the Unitree Aliengo using a trotting gait at 0.2 m/s; higher speeds caused instability and falls. Applicability to other platforms, gaits, and velocities remains to be demonstrated.

\section{CONCLUSIONS AND FUTURE WORK}
This work presented a proprioceptive terrain mapping framework for quadruped robots to support exploration in permanently shadowed regions of the Moon. The approach generates a multi-layer 2.5D gridmap from internal sensing, encoding elevation, slippage, energetic cost, and gravito-inertial stability margins, offering a robot-centered description of terrain interaction that complements exteroceptive perception in low-visibility environments.

Lunar simulations showed that the proposed metrics captured spatial patterns aligned with terrain geometry, including localized slippage at slope transitions. The energetic cost varied non-monotonically with inclination due to controller adaptation, and stability margins decreased with slope magnitude. Mapping multiple routes revealed segments of elevated energetic demand or reduced stability, supporting path comparison beyond global averages.

These results indicate that proprioceptive mapping can enhance terrain assessment in PSRs by linking the robot’s physical interaction with the environment to decision-making. Future work will validate the framework’s physical fidelity through comparisons with terramechanics models, alternative physics engines, and lunar analog tests, while integrating proprioceptive maps with exteroceptive and semantic data to improve robustness and situational awareness in future missions.

\bibliographystyle{IEEEtran}
\bibliography{main}

\end{document}